\documentclass{ieeeaccess}
\usepackage{cite}
\usepackage{amsmath,amssymb,amsfonts}
\usepackage{algorithmic}
\usepackage{graphicx}
\usepackage{textcomp}
\usepackage[colorlinks=true, citecolor=blue, linkcolor=blue, urlcolor=blue, breaklinks=true]{hyperref}

\usepackage{bm}
\makeatletter
\AtBeginDocument{\DeclareMathVersion{bold}
\SetSymbolFont{operators}{bold}{T1}{times}{b}{n}
\SetSymbolFont{NewLetters}{bold}{T1}{times}{b}{it}
\SetMathAlphabet{\mathrm}{bold}{T1}{times}{b}{n}
\SetMathAlphabet{\mathit}{bold}{T1}{times}{b}{it}
\SetMathAlphabet{\mathbf}{bold}{T1}{times}{b}{n}
\SetMathAlphabet{\mathtt}{bold}{OT1}{pcr}{b}{n}
\SetSymbolFont{symbols}{bold}{OMS}{cmsy}{b}{n}
\renewcommand\boldmath{\@nomath\boldmath\mathversion{bold}}}
\makeatother

\def\BibTeX{{\rm B\kern-.05em{\sc i\kern-.025em b}\kern-.08em
    T\kern-.1667em\lower.7ex\hbox{E}\kern-.125emX}}

\begin{document}
\history{Date of publication xxxx 00, 0000, date of current version xxxx 00, 0000.}
\doi{10.1109/ACCESS.2024.0429000}

\title{Wildfire spread forecasting with Deep Learning}
 
\author{\uppercase{Nikolaos Anastasiou}\authorrefmark{1},
\uppercase{Spyros Kondylatos}\authorrefmark{1, 2}, and \uppercase{Ioannis Papoutsis}\authorrefmark{1,3}}

\address[1]{Orion Lab, National Observatory of Athens \& National Technical University of Athens}
\address[2]{Image Processing Laboratory (IPL), Universitat de Val\`encia}
\address[3]{Archimedes, Athena Research Center}

\tfootnote{This work was funded by the project: Small-Satellites (Mea-
sure ID 16855), implemented by the Hellenic Ministry of
Digital Governance with the European Space Agency (ESA)
Assistance in the Management and Implementation.}

\markboth
{Anastasiou \headeretal: Wildfire spread forecasting with deep learning}
{Anastasiou \headeretal: Wildfire spread forecasting with deep learning}

\corresp{Corresponding author: Nikolaos Anastasiou (e-mail: rs20095@mail.ntua.gr).}

\begin{abstract}
Accurate prediction of wildfire spread is crucial for effective risk management, emergency response, and strategic resource allocation. In this study, we present a deep learning (DL)-based framework for forecasting the final extent of burned areas, using data available at the time of ignition. We leverage a spatio-temporal dataset that covers the Mediterranean region from 2006 to 2022, incorporating remote sensing data, meteorological observations, vegetation maps, land cover classifications, anthropogenic factors, topography data, and thermal anomalies. To evaluate the influence of temporal context, we conduct an ablation study examining how the inclusion of pre- and post-ignition data affects model performance, benchmarking the temporal-aware DL models against a baseline trained exclusively on ignition-day inputs. Our results indicate that multi-day observational data substantially improve predictive accuracy. Particularly, the best-performing model, incorporating a temporal window of four days before to five days after ignition, improves both the F1 score and the Intersection over Union by almost 5\% in comparison to the baseline on the test dataset. We publicly release our dataset and models to enhance research into data-driven approaches for wildfire modeling and response.
\end{abstract}


\titlepgskip=-21pt

\maketitle
\section{Introduction}
\label{sec:introduction}
\PARstart{W}{ildfires} are natural disturbances that play a crucial role in maintaining ecosystem health and biodiversity, offering important ecological benefits \cite{https://doi.org/10.1002/fee.2044}.
Nevertheless, they also pose significant threats to human life, infrastructure \cite{kramer2019wildfire}, and public health, due to the release of particulate matter and toxic pollutants \cite{Finlay2012, {https://doi.org/10.1155/2014/958457}}. 
With climate change projected to intensify both the frequency and severity of wildfires \cite{jones2020climate}, the development of resilient systems for enhancing wildfire management has become increasingly urgent.

One of the most critical wildfire-related applications is the forecasting of wildfire spread, which aims to predict the future evolution of a wildfire following its ignition \cite{valero2017integrated}.
The implementation of fire spread models varies considerably between studies, particularly in terms of the spatial and temporal resolution.
Thus, applications range from near-real-time hourly forecasts\cite{mandel2014recent} to cumulative estimates of the total burned area throughout an entire fire event\cite{marjani2024cnn}, and even to early predictions of final fire extent immediately after ignition\cite{alizadeh2024fusionfirenet}.

Satellite imagery, when combined with weather measurements, can play a vital role in identifying regions at high risk of ignition and spread\cite{richardson2022global}. 
However, the accuracy of these predictions is constrained by the spatial resolution of meteorological forecasts, along with the revisit frequency and coverage of satellite sensors \cite{shadrin2024wildfire}.
As a result, unmanned aerial vehicles (UAVs) are often employed for high-resolution fire monitoring\cite{afghah2019wildfire}, whereas satellites are better suited for broader, multi-day forecasts across large geographic areas\cite{huot2022next}.

Wildfire spread is driven by complex, non-linear interactions between atmospheric conditions, fuel properties (such as vegetation type and density), and topography. 
Wind, in particular, has a significant effect on the direction and extent of burned areas \cite{pausas2021wildfires}. 
Historically, wildfire spread prediction has been estimated using empirical systems like the Canadian Forest Fire Behavior Prediction System\cite{hirsch1996canadian}. 
More recent physics-based models, such as WRF-SFIRE\cite{mandel2014recent}, have enabled near real-time burned area forecasts\cite{kale2024operational}, relying on high-resolution meteorological inputs.

Lately, advances in Machine Learning (ML) and Deep Learning (DL) have opened new avenues for data-driven approaches to wildfire modeling \cite{reichstein_deep_2019}, showing great promise in predicting wildfire spread \cite{radke2019firecast}.
These methods can be trained in large-scale datasets---including satellite imagery, meteorological records, and historical fire perimeters---to learn patterns and capture the complex spatio-temporal behavior of wildfires.

Motivated by recent advances, this study explores DL-based approaches for capturing the spatiotemporal context surrounding wildfires to predict their spread, specifically aiming to estimate the final extent of burned areas starting from an ignition point.
The proposed approach leverages satellite imagery, meteorological data, and topographic information.
We explore both Vision Transformers (ViTs) and Convolutional Neural Networks (CNNs)-based architecture using input features from the ignition and preceding days, as well as information for the post-ignition period. 
The main contributions of this work can be summarized as follows:
\begin{itemize}
\item Building on the Mesogeos datacube~\cite{kondylatos2023mesogeos}, we develop a novel, large-scale ML-ready dataset comprising approximately 9,500 fire events, containing spatio-temporal information of fire drivers and curated for wildfire spread analysis. The dataset is designed to be readily applicable for training and evaluating data-driven models.
\item We train DL models that predict the final burned area, using the available state of relevant wildfire-driving variables at the time of the ignition. Additionally, we provide a quantitative and qualitative evaluation of model predictions, including visual comparisons of predicted and actual burned areas for selected events to assess the practical utility of the predictions.
\item We perform an ablation study to assess the effect of varying temporal input windows on model performance. Our findings demonstrate that increasing both pre- and post-ignition information context on the input variables significantly enhances predictive accuracy.
\end{itemize}

To promote reproducibility and facilitate future research in this critical area, the dataset and source code for the developed models are being made publicly available at \url{https://github.com/Orion-AI-Lab/WildFireSpread}.

\section{Related Work} 

DL techniques have been increasingly applied in a range of wildfire-related applications, including wildfire danger forecasting, active fire detection, burned area mapping, and fire spread prediction \cite{jain_review_2020}.
Le et al \cite{le_new_2021} used a Multilayer Perceptron for predicting wildfire danger, while other studies use CNNs to model forest fire susceptibility \cite{zhang_forest_2019, zhang_deep_2021, bergado_predicting_2021, bjanes_deep_2021}.
Kondylatos et al. \cite{kondylatos2022wildfire} used models like Long Short-Term Memory (LSTM) and Convolutional LSTM (ConvLSTM), demonstrating improved performance over traditional indices such as the Fire Weather Index (FWI) \cite{van1974structure}, and used explainable Artificial Intelligence to uncover the main drivers that influence the occurrence of fires.

CNNs have also proven effective for the rapid detection of active wildfires from various sources, including aerial vehicle thermal imagery \cite{bouguettaya2022review, ghali2022deep} and satellite images, which cover a broader spatial extent \cite{toan2019deep} with frequent data updates \cite{aminou2002msg}.
Recently, a transformer-based solution was proposed to segment active fire pixels from the VIIRS satellite, demonstrating improved performance compared to other spatial and temporal-aware models \cite{zhao_tokenized_2023}.
Moreover, DL models have accurately segmented burned areas from high-resolution multispectral imagery (e.g., Sentinel-2), often outperforming traditional methods like Normalized Burn Ratio (NBR)\cite{escuin2008fire} thresholding, especially in handling varying burn severities \cite{knopp2020deep, sdraka2024floga}.

\textbf{Wildfire Spread}

Recent studies have explored a variety of DL approaches for modeling wildfire spread, with a strong emphasis on capturing the complex spatial and temporal dynamics involved. Shadrin et al. \cite{shadrin2024wildfire} evaluated several CNN-based architectures for wildfire spread prediction, emphasizing the effectiveness of convolutional models in learning spatial patterns from geospatial data.
Large-scale datasets such as Next Day Wildfire Spread \cite{huot2022next} and WildfireSpreadTS \cite{NEURIPS2023_ebd54517} have been introduced, enabling daily prediction of wildfire spread and benchmarking DL models in this task. 
FireCast \cite{radke_firecast_2019} used CNNs trained on historical fire data to identify high-risk zones for imminent spread. Burge et al. \cite{burge_convolutional_2021} applied a 
Convolutional Long Short-Term Memory (ConvLSTM) network to model wildfire propagation dynamics using simulated datasets. Hodges and Lattimer \cite{hodges_wildland_2019} proposed a Deep Convolutional Inverse Graphics Network to predict fire spread up to six hours in advance.
Marjani et al. \cite{marjani2024cnn} combined convolutional layers with Bidirectional Long Short-Term Memory units (CNN-BiLSTM), effectively capturing spatio-temporal dependencies in wildfire spread data. 
Coffield et al. \cite{coffield2019machine} employed traditional ML models, including decision trees, to estimate final fire size at the time of ignition. 
More recently, Kondylatos et al. \cite{kondylatos2023mesogeos} addressed the same problem using CNNs, relying solely on input features from the day of ignition.
While the majority of prior work has emphasized spatial extent prediction, some studies focus on estimating the final numerical scale of wildfires. For example, Liang et al. \cite{liang2019neural} employed an LSTM model with meteorological inputs to classify final fire sizes into categorical levels, underscoring the critical influence of weather conditions on fire outcomes.

Despite these advancements, to the best of the authors' knowledge, no study has systematically investigated the optimal temporal window of input data, encompassing both pre- and post-ignition periods, for this task, particularly using advanced DL architectures and large-scale, multi-year spatio-temporal datasets.
This work aims to address this gap by exploring the integration of ViT and CNN-based models with temporally-structured RS and meteorological data to predict final fire extent at the ignition stage.

\section{Dataset}
Training DL models for wildfire spread prediction requires high-quality, high-resolution data. 
For this study, a dedicated dataset was created by extracting and processing data from Mesogeos \cite{kondylatos2023mesogeos}, a large-scale multi-purpose dataset for wildfire modeling in the Mediterranean.

\subsection{Variables}
\label{sec:variables}
Mesogeos comprises 27 variables at a spatial resolution of $1 \times 1$ km and a daily temporal resolution spanning the period from 2006 to 2022. 
It includes 14 dynamic (time-varying) and 14 static (time-invariant) features, encompassing meteorological parameters, land cover, vegetation indices, and topographical characteristics. 
The variables selected for this study were chosen based on their established relevance to wildfire ignition and spread (Table \ref{tab4} in Appendix).

Weather conditions play a crucial role in determining the rates of fuel drying, directly affecting fire occurrence and spread.
The dataset includes key atmospheric variables such as wind speed and direction, air temperature (at 2 meters), dew point temperature (at 2 meters), relative humidity, surface pressure, total precipitation (24-hour), and surface solar radiation downwards.
Among these, wind speed and direction are particularly critical, as fire spread is highly sensitive to wind dynamics.
Relative humidity provides a proxy for fuel moisture content, while temperature and solar radiation affect both fuel drying and fire intensity. 
High precipitation can play a mitigating role by reducing fuel flammability and limiting fire propagation.

To enhance the dataset's suitability for fire spread modeling, we derived some additional variables. 
In particular, wind speed was resolved into its eastward (U) and northward (V) vector components, enabling a more precise representation of wind-driven fire behavior.

\[
U = W \cos(\theta)
\]
\[
V = W \sin(\theta)
\]
where:
\begin{itemize}
    \item \( W \) is the Wind Speed (in m/s),
    \item \( \theta \) is the Wind Direction (in degrees).
\end{itemize}


Vegetation characteristics are key indicators of fuel type and availability. 
The dataset includes variables such as the Normalized Difference Vegetation Index (NDVI), Leaf Area Index (LAI), and surface soil moisture. 
High NDVI and LAI values signal dense vegetation and greater fuel loads, while low soil moisture increases fuel flammability, promoting easier ignition and more intense burning.


Static variables describing the land surface and terrain also play a crucial role in shaping fire spread dynamics. 
Land cover classification identifies dominant vegetation types and surface features, such as agricultural land, forests, grasslands, shrubland, sparse vegetation, settlements, and water bodies---each with distinct flammability and fire propagation characteristics.
For instance, settlements and water bodies often act as natural barriers to the spread. 
Topographical features, including elevation, slope, aspect, and curvature, also play a vital role, with fires typically spreading more rapidly uphill due to convective heat and terrain-driven wind patterns.

Two key binary variables are included for each sample: the ignition point and the final burned area mask. The ignition point is encoded as a binary image, with a value of 1 at the pixel where the fire originated and 0 elsewhere. The burned area mask serves as the prediction target, represented as a binary image in which burned pixels are labeled as 1 and unburned pixels as 0.


\subsection{Study Area}
The dataset covers the Mediterranean region, spanning an area of approximately $3,760,000\ km^2 $. 
This region is of particular interest for wildfire research due to its characteristic Mediterranean climate, characterized by hot and dry summers and a high frequency of fire activity \cite{nastos2009mediterranean}.
Wildfires in this region contribute significantly to carbon emissions \cite{garcia2013carbon}, and are expected to become more frequent and severe under ongoing climate change \cite{ruffault2020increased}. 
The dataset includes fire events from multiple countries within this region, capturing a wide range of landscapes and climatic sub-regions. 
This geographic diversity enables the analysis of cross-border fire patterns beyond those possible with single-country datasets.

\Figure[h](topskip=0pt, botskip=0pt, midskip=0pt)[width=0.98\linewidth]{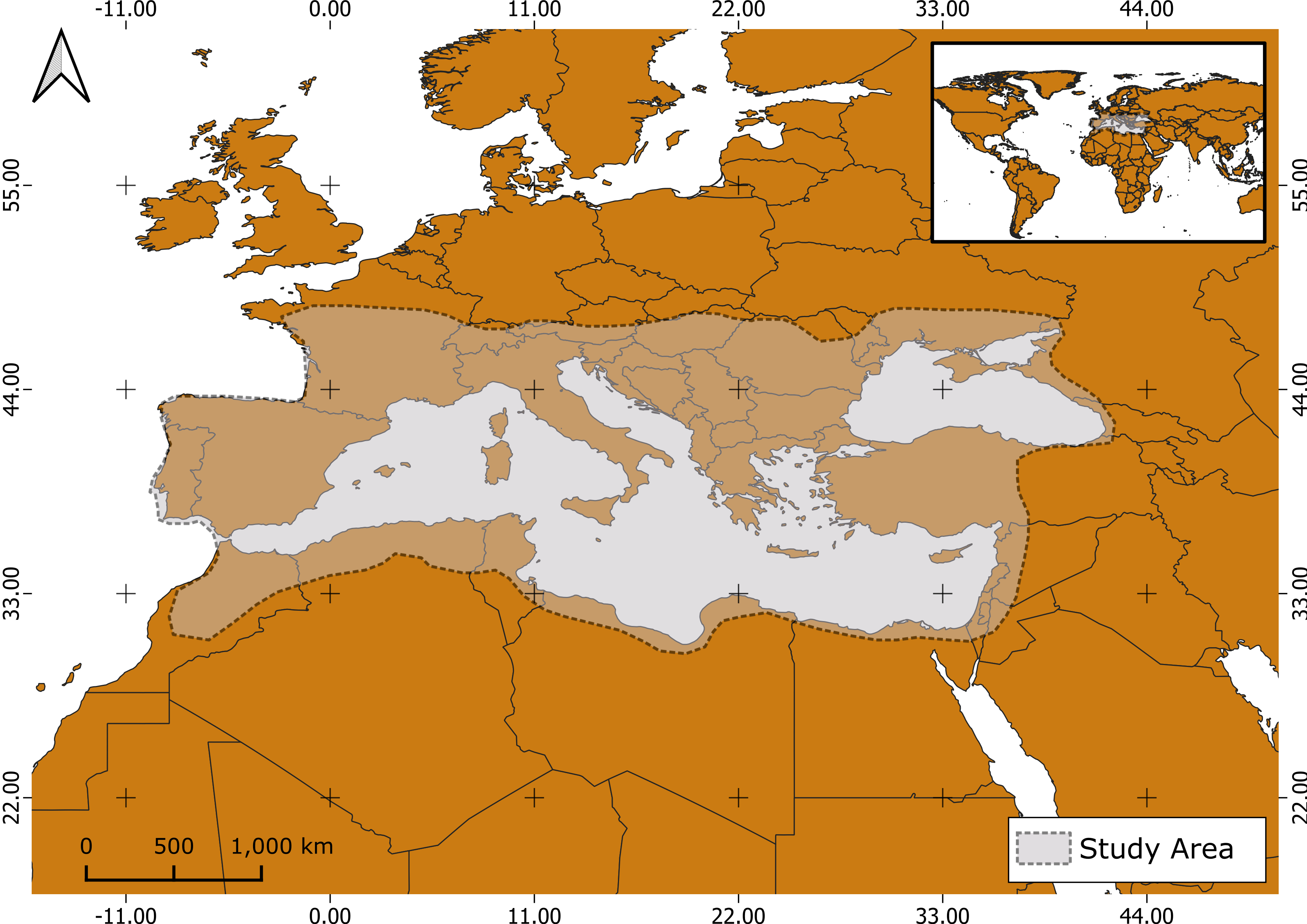}
{ \textbf{Location of the study area (Coordinate System: WGS84).}\label{fig1}}

\subsection{Dataset Extraction and Preprocessing}
For this study, we derived an ML–ready dataset from the Mesogeos datacube. 
The dataset was generated by extracting spatio-temporal patches centered around historical wildfire ignition points. 
Specifically, for each fire event, a $64 \times 64$ pixel ($64 \times 64$ km) patch was extracted, incorporating all the variables mentioned in \ref{sec:variables}.
To mitigate potential spatial bias and avoid always centering the burned area, a small random offset was applied to the patch center relative to the recorded ignition point.

To capture the temporal dynamics relevant to wildfire spread, data were extracted over a temporal window spanning from four days before the ignition day to five days after. 
Longer-term forecasts are constrained by the current accuracy of meteorological inputs.
As a result, each sample (fire event) is a spatio-temporal datacube with dimensions of (time steps $\times$ height $\times$ width) = (10 $\times$ 64 $\times$ 64).

Several post-processing steps were conducted to ensure data integrity. 
Missing values—primarily due to ERA5-LAND’s land-only coverage affecting areas over water—were imputed with zeros. 
Samples with substantial coverage over water bodies were excluded to maintain focus on terrestrial wildfire behavior. 
The final dataset comprises 9,568 wildfire events.

\subsection{Dataset Statistics}
To provide insights into the dataset composition, several descriptive statistics were computed based on the 9,568 total samples.

Figure \ref{fig2} displays the annual distribution of fire events from 2006 to 2022. 
An upward trend is observed in recent years, particularly after 2017 (excluding 2018), where the number of recorded events exceeds 650 per year. In contrast, earlier years contain significantly fewer samples.

\Figure[h!](topskip=0pt, botskip=0pt, midskip=0pt)[width=0.98\linewidth]{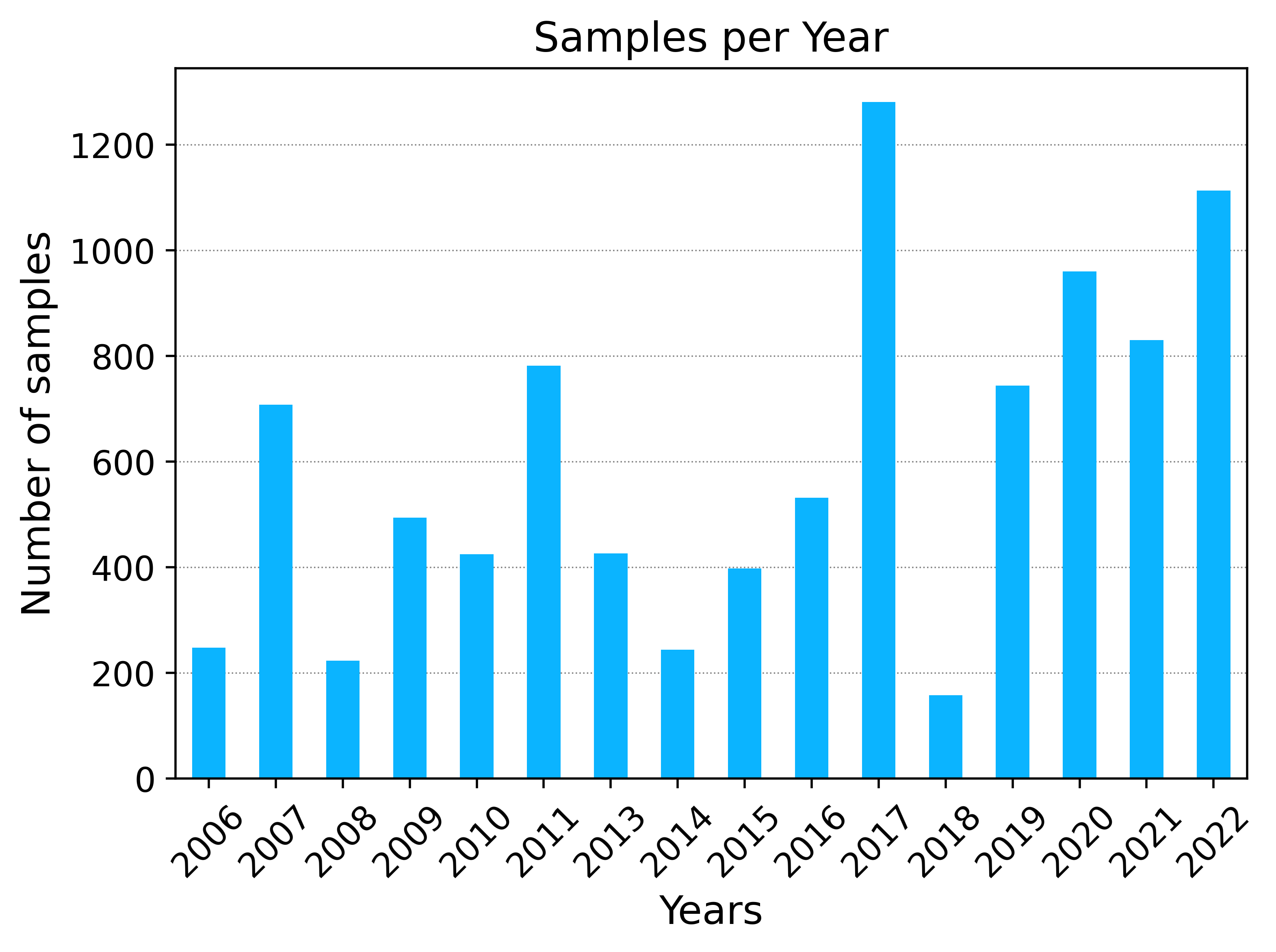}
{ \textbf{Yearly distribution of fire events in the dataset.}\label{fig2}} 

Figure \ref{fig3} presents the cumulative monthly distribution of fire events aggregated across all years. 
As expected in a Mediterranean climate, the majority of fires occur during the summer months, with July and August exhibiting the highest frequency, followed by March and September, though at notably fewer occurrences.

\Figure[h!](topskip=0pt, botskip=0pt, midskip=0pt)[width=0.98\linewidth]{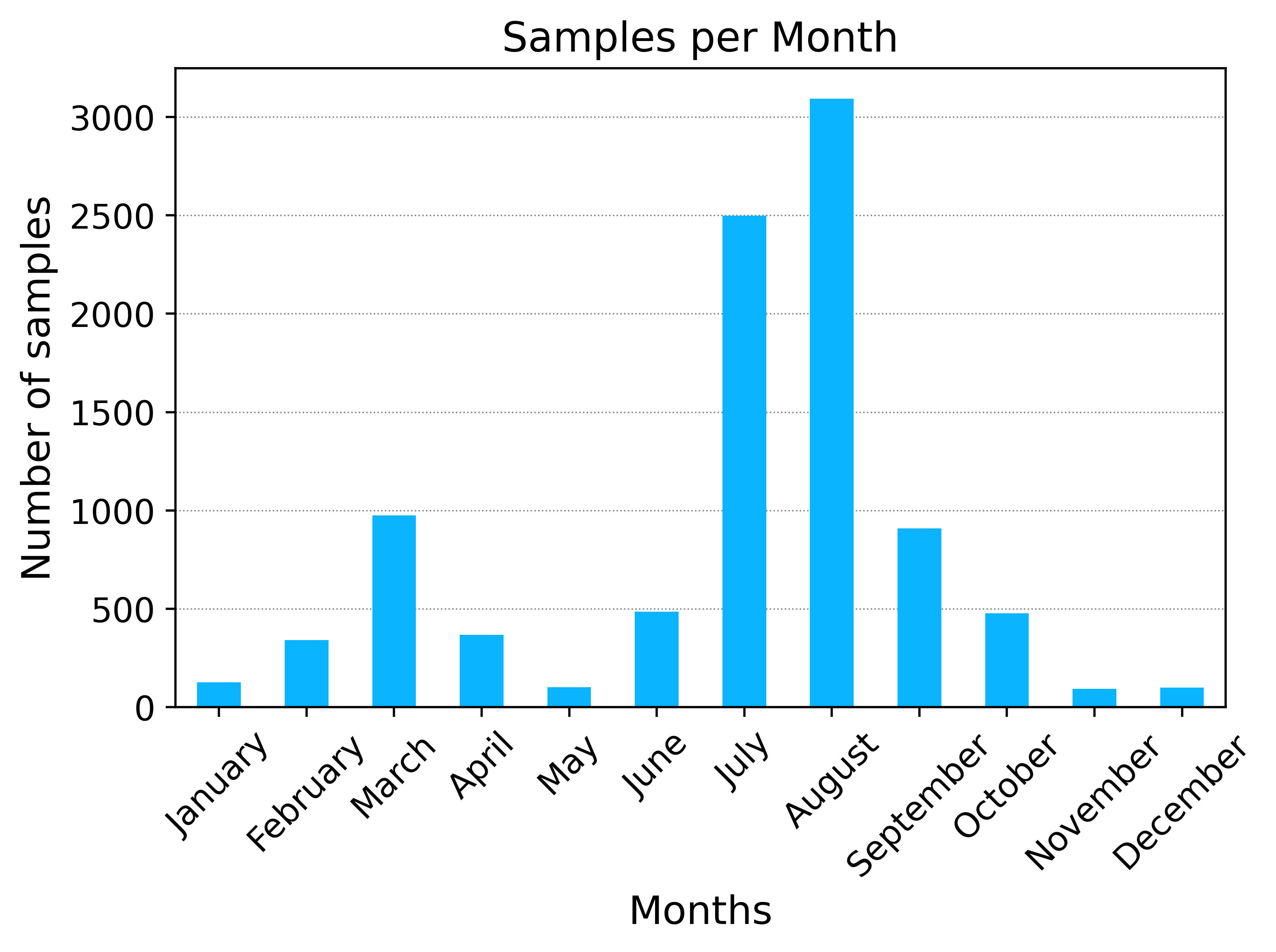}
{ \textbf{Monthly distribution of fire events across all years in the dataset.}\label{fig3}} 

To assess the variability in wildfire sizes, fire events were grouped using k-means clustering \cite{likas2003global} based on their final burned area.
An optimal cluster count of $k=10$ was selected to adequately capture the distribution of fire sizes. 
Figure \ref{fig4} presents the number of samples in each cluster, with ranges defined by the minimum and maximum burned area (in hectares) per cluster.
The results reveal a significant class imbalance, with the majority of fire events concentrated in the smallest size cluster (<1500 ha), while in contrast, larger fires are underrepresented. 
This class imbalance is a known challenge in data-driven wildfire modeling \cite{prapas2021deeplearningmethodsdaily}. 

\Figure[h!](topskip=0pt, botskip=0pt, midskip=0pt)[width=0.98\linewidth]{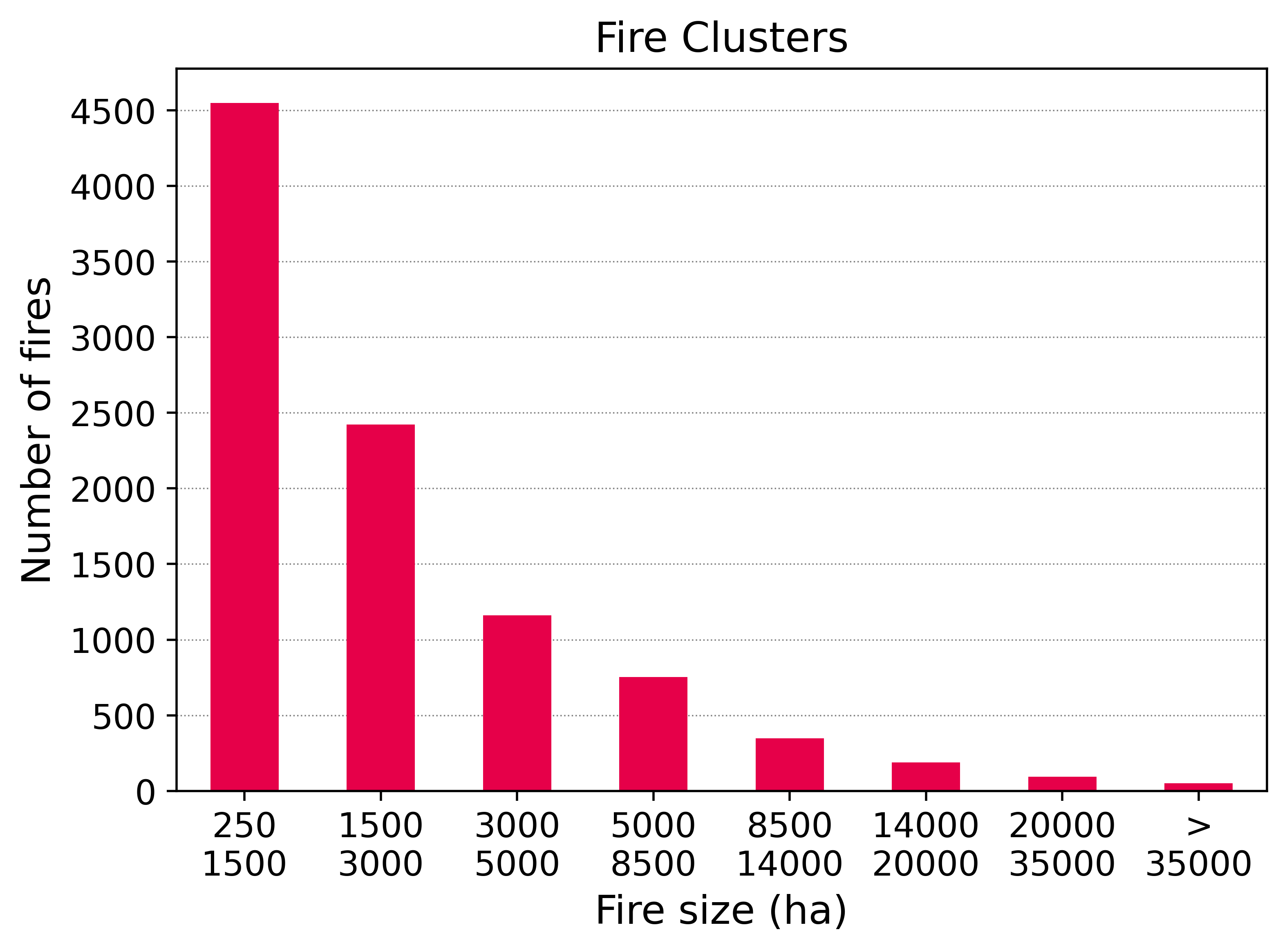}
{ \textbf{Number of samples (fire events) per fire cluster in the dataset, represented in hectares (ha).}\label{fig4}}

\newpage
\section{Methodology}
This section first details the experimental setup, followed by a description of the models developed, and finally outlines the evaluation procedure followed in this study.

\subsection{Experimental Setup}

Each data sample comprises 27 variables, including both static and dynamic features. 
The final model input consists of 14 dynamic variables (excluding burned areas), each recorded over 10 time steps, and 12 static variables, including the ignition point, loaded once per sample. 
The 10 time steps cover a temporal window from 4 days before to 5 days following the fire ignition, forming a time series. 
Figure \ref{fig5} illustrates the structure of the data for a single fire event (sample). 
All variable data are normalized to the range $[0, 1]$.

The final burned area variable serves as the prediction target (label) for all models. 
The model output is a binary segmentation mask with dimensions identical to the input channels ($64 \times 64$ pixels), where burned pixels are labeled as 1 and unburned pixels as 0. 
This mask represents the predicted final burned area resulting from a wildfire event, as shown in Figure \ref{fig6}.
The dataset exhibits significant class imbalance between burned and unburned pixels; within each $64 \times 64$ patch ($4096$ pixels in total), on average, only about 100 pixels correspond to burned areas, with the remainder being unburned.

For training and evaluation, we adopted a temporal split. 
Fire events from 2006 to 2020 were allocated to the training set, the year 2021 was used for validation during model training to monitor performance and tune hyperparameters, and the year 2022 was reserved as the test set. 
This resulted in 7619 fire events for training, 841 for validation, and 1101 for testing.


\Figure[h!](topskip=0pt, botskip=0pt, midskip=0pt)[width=0.98\linewidth]{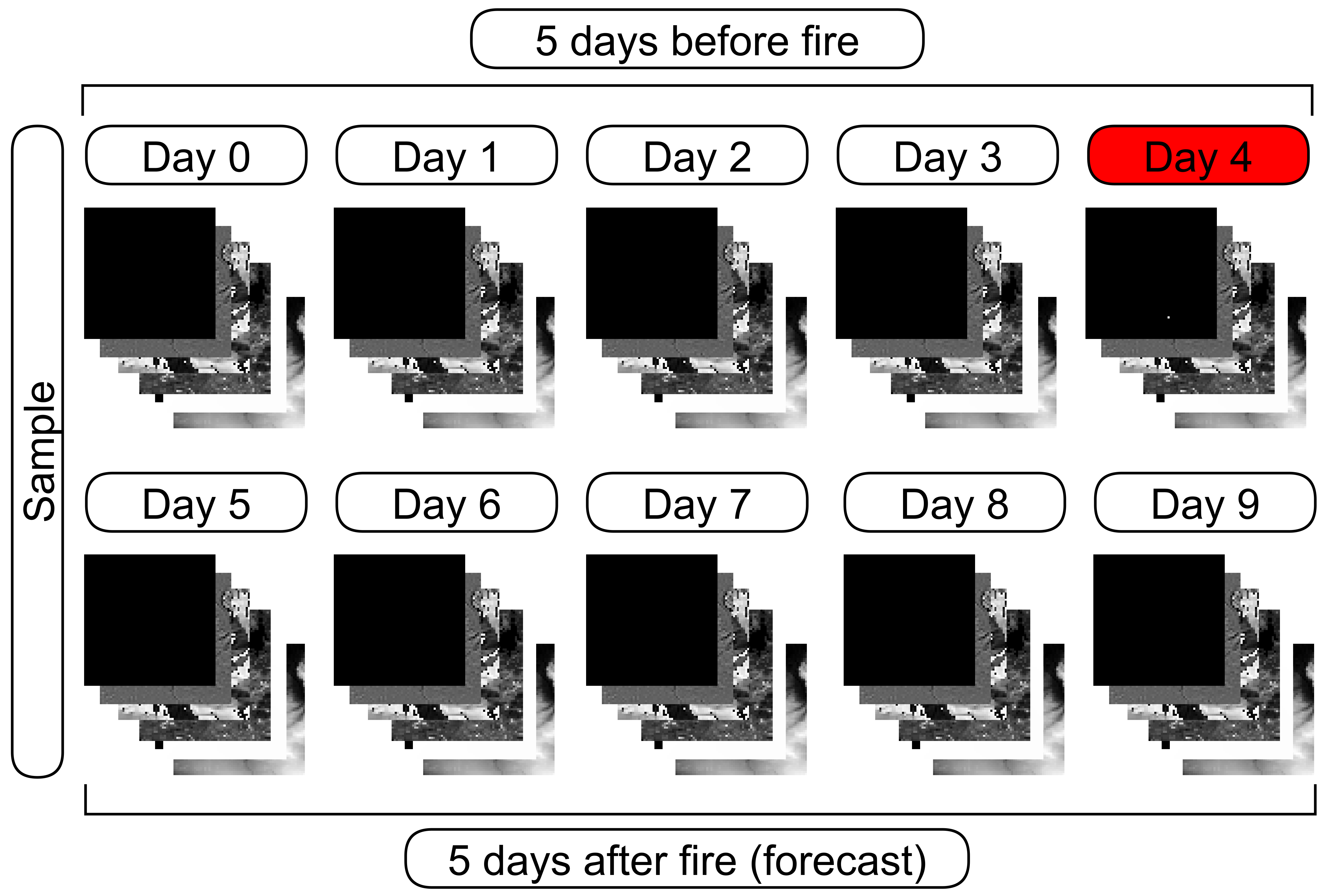}
{ \textbf{Fire event (sample) structure, containing 4 days before fire ignition day, day 4 (ignition day), and five days after ignition day. Each day incorporates 27 variables, including both dynamic and static variables.}\label{fig5}} 

\Figure[h!](topskip=0pt, botskip=0pt, midskip=0pt)[width=0.99\linewidth]{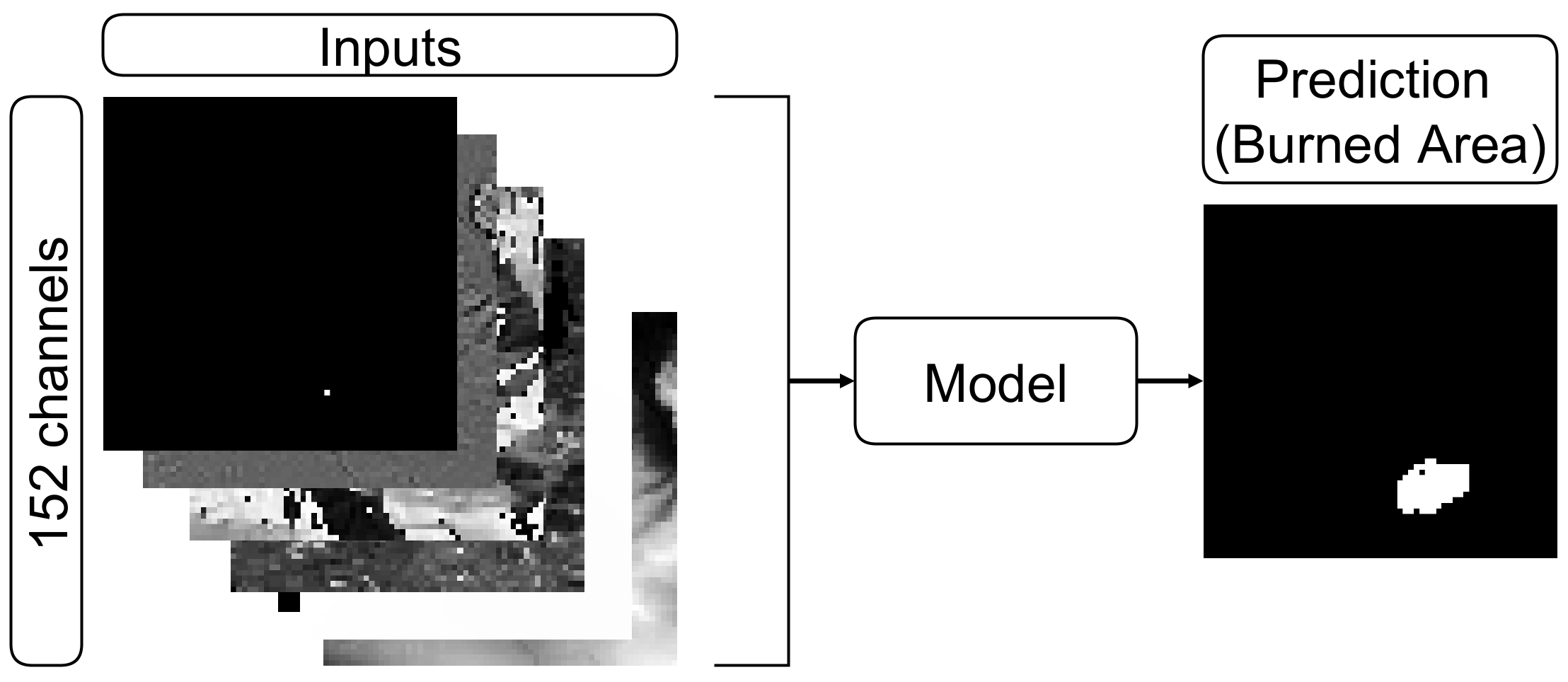}
{ \textbf{The model takes as input 152 channels (including the ignition point) representing a 10-day temporal window, and produces a 64 $\times$ 64 pixels binary masks, where burned pixels are labeled as 1 and unburned pixels as 0.}\label{fig6}}

\subsection{Algorithms}
CNNs are well-suited algorithms for extracting hierarchical spatial features, making them a natural choice for modeling wildfire spread patterns. 
However, standard 2D CNNs process each variable at each time step independently, limiting their ability to directly capture temporal dependencies inherent in fire spread dynamics.
To address this limitation, each input sample was structured as a multi-channel image, where each channel corresponds to a specific variable at a given time step. 
This design effectively makes use of temporal information as additional input dimensions for 2D CNN models.
The CNN models in this study were based on the U-Net\cite{ronneberger2015unetconvolutionalnetworksbiomedical} architecture.

First, a baseline model was implemented using a modified 2D U-Net architecture trained exclusively on data from the ignition day. 
This model utilized 26 input channels (excluding the burned area label) and aimed to establish the predictive capability achievable with minimal temporal context.

Second, a principal 2D U-Net model was trained using the full 10-day temporal window. 
This input comprised 152 channels (14 dynamic variables $\times$ 10 time steps $+$ 12 static variables). 
The inclusion of pre-ignition data allows the model could learn pre-fire environmental conditions indicative of fire spread risk, while post-ignition data capture critical changes in dynamic factors, such as wind patterns, that directly influence the final burned area extent and direction.

Finally, a modified 3D U-Net\cite{wang2023review} architecture was trained using the same spatio-temporal dataset. 
Unlike the 2D U-Net, the 3D U-Net employs 3D convolutional filters that simultaneously process spatial and temporal dimensions. 
This architecture inherently models temporal dependencies more directly and effectively than the 2D channel-stacking approach, enabling improved feature learning across consecutive time steps.

Figure \ref{fig7} presents the U-Net architecture employed in this study which was modified from the original U-Net design to better accommodate the characteristics of the input data. 
Specifically, two encoder and decoder blocks were removed resulting in a shallower network architecture, while an extra convolutional layer was added within each convolutional block to enhance local feature representation as shown Figure \ref{fig8}.

This architectural adjustment was motivated by the relatively low spatial resolution of the input samples ($64 \times 64$ pixels), which does not require the extensive hierarchical feature extraction provided by a full U-Net with numerous pooling layers. 
Instead, the primary complexity in this task arises primarily in capturing patterns between the numerous temporal and variable channels. 

Another key architectural modification was the replacement of the standard Rectified Linear Unit (ReLU)\cite{he2018relu} activation function with the Gaussian Error Linear Unit (GELU)\cite{hendrycks2016gaussian}, which was applied in both the 2D and 3D U-Net models. 
Additionally, the models were trained using the BCEDice Loss function\cite{galdran2022optimal}, which combines Binary Cross-Entropy and the Dice Loss, allowing the models to classify pixels as burned or unburned while simultaneously optimizing for spatial context.

Apart from the U-Net, we also experimented with a ViT model architecture to explore the potential of attention-based mechanisms for this task.

\Figure[h!](topskip=0pt, botskip=0pt, midskip=0pt)[width=0.9\linewidth]{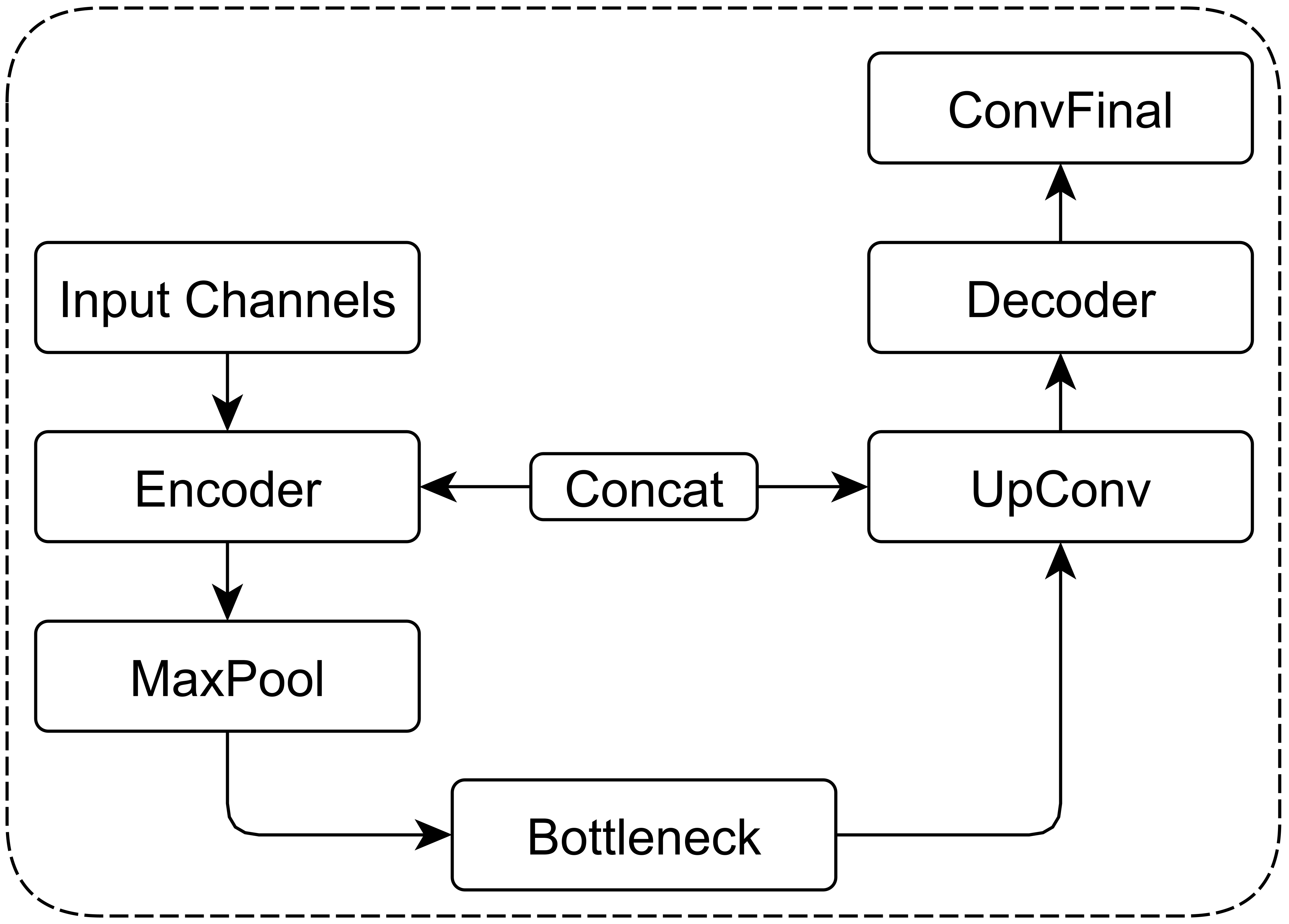}
{ \textbf{The modified architecture for both U-Net2D and U-Net3D consists of an Encoder Block followed by a MaxPooling Layer and a Bottleneck. This is followed by an UpConvolution Layer, which is concatenated with the corresponding Encoder Block, and a Decoder Block, culminating in a final Convolution Layer. The Encoder, Bottleneck, UpConvolution, and Decoder Blocks each include a Convolution Block, which is illustrated in Fig. \ref{fig8}.\label{fig7}} }

\Figure[h!](topskip=0pt, botskip=0pt, midskip=0pt)[width=0.9\linewidth]{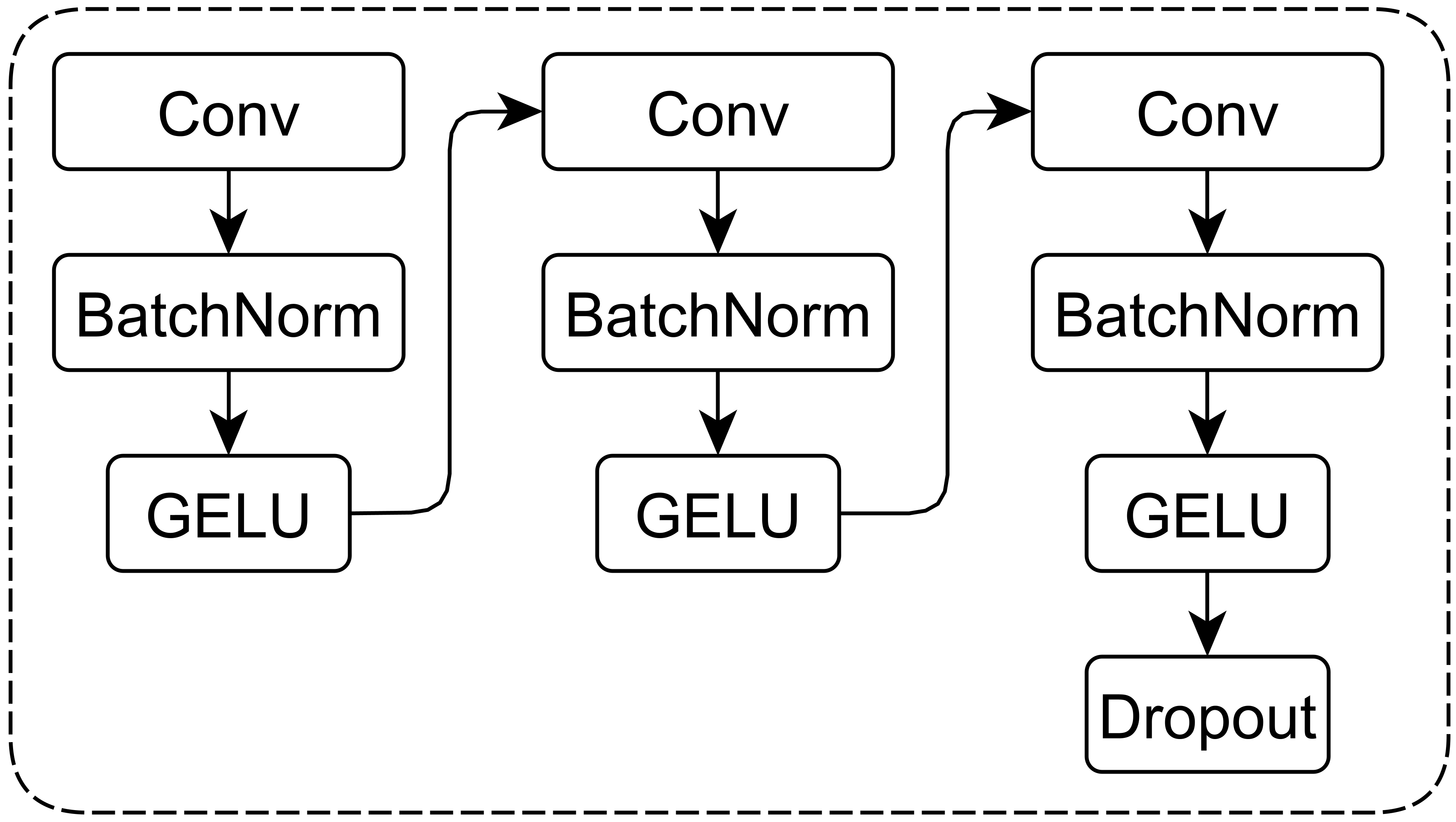}
{ \textbf{The modified Convolution Block used in both U-Net2D and U-Net3D models consists of a Convolution Layer followed by Batch Normalization and a GELU (Gaussian Error Linear Unit) Activation function. This sequence is repeated twice, with a Dropout Layer added after the second repetition.}\label{fig8}}

\subsection{Evaluation Metrics}
Model performance was evaluated using standard evaluation metrics commonly applied in binary segmentation tasks: Precision, Recall, Dice Score, F1-Score, and Intersection over Union (IoU). 
Notably, in the context of binary 2D segmentation, the F1-score is mathematically equivalent to the Dice Score. 
As such, only the Dice Score is included in the results tables for clarity.
Among these scores, the Dice Score was selected as the primary criterion for hyperparameter tuning and model selection during training, due to its effectiveness in quantifying the spatial overlap between predicted and reference burned areas.
While IoU is also reported, it serves as a more conservative metric, placing greater emphasis on penalizing both false positives and false negatives. 
The mathematical definitions of all evaluation metrics are provided below:

\begin{flushright}
\[
Precision = \frac{True Positives}{True Positives + False Positives}, \quad [0,1]
\]

\[
Recall = \frac{True Positives}{True Positives + False Negatives}, \quad [0,1]
\]

\[
F1Score = 2 \cdot \frac{Precision \cdot Recall}{Precision + Recall}, \quad [0,1]
\]

\[
Dice Score = \frac{2 \cdot TP}{2 \cdot TP + FP + FN} = \frac{2 |A \cap B|}{|A| + |B|}, \quad  [0,1]
\]

\[
IoU = \frac{TP}{TP + FP + FN} = \frac{|A \cap B|}{|A \cup B|}, \quad [0,1]
\]
\end{flushright}

\section{Results}
The evaluation of model performance is presented in two primary sections to highlight both overall effectiveness and the influence of key factors on model performance.
First, we report the aggregate performance metrics, calculated over all 1,101 test fire events, to provide a general assessment of model accuracy.
However, these metrics alone may not fully capture performance variations related to the different fire sizes.
To provide deeper insight into model performance, we also report results by fire size clusters, distinguishing between small, medium, and large events.
Furthermore, we present results from a series of models trained with varying temporal windows.
This analysis aims to identify the optimal temporal configuration and assess the extent to which incorporating additional temporal context improves the model's ability to predict wildfire spread.

\subsection{Models Comparison}
Table~\ref{tab1} summarizes the overall performance of the four evaluated models on the test set: the baseline ignition-day U-Net2D, the 10-day U-Net2D, the 10-day U-Net3D, and the 10-day ViT.
The results demonstrate that the 10-day U-Net3D model achieved the highest performance, outperforming both the baseline and the 10-day U-Net2D variants.
In contrast, the ViT model exhibited the lowest performance, which may be attributed to the limited dataset size relative to the data-intensive nature of transformer-based architectures.

\begin{table}[h!]
\centering
\caption{\textbf{Evaluation results of the models on the test set. Higher performance metrics are indicated in bold.}}
\label{tab1}
\setlength{\tabcolsep}{8pt}
\renewcommand{\arraystretch}{1.2}
\begin{tabular}{ccccc}
\hline
\textbf{Metric} & \textbf{U-Net2D} & \textbf{U-Net2D} & \textbf{U-Net3D} & \textbf{ViT} \\
\textbf{\%}    & \textbf{Baseline} & \textbf{10 days} & \textbf{10 days} & \textbf{10 days} \\
\hline
Dice Score & 48.3 & 51.7 & \textbf{53.6} & 43.7 \\
IoU        & 31.9 & 34.8 & \textbf{36.6} & 28.0 \\
Precision  & 50.2 & 58.6 & \textbf{59.6} & 59.4 \\
Recall     & 46.6 & 46.2 & \textbf{48.8} & 34.6 \\
\hline
\end{tabular}
\end{table}

The top-performing model achieved a Dice Score of 53.6\% on the test set. 
The corresponding IoU was 36.6\%, reflecting the stringent nature of the IoU metric in penalizing segmentation errors. 
Analyzing the components of the F1-Score, the model demonstrated a Precision of 59.6\%, suggesting a relatively low rate of false positives and a conservative prediction of the burned area extent as evident from the low Recall rate of 48.8\%. Such conservatism contributes to reducing the risk of overestimating the final burned area, while potentially underrepresenting some affected regions.
 
The strong performance of the DL models highlights the significant potential of data-driven approaches for modeling wildfire spread.
This effectiveness can be attributed to the ability of DL models to learn from the complex, nonlinear interactions among the multiple fire drivers---such as vegetation types, land cover, topographic features, and meteorological observations.
By jointly modeling these diverse inputs, DL methods offer a robust framework for capturing the multifaceted dynamics that govern fire behavior.

\subsection{Influence of the temporal window}

One main part of this study is to assess the impact of the temporal window size, particularly the extent of post-ignition data.
For this, additional experiments are conducted by training a series of models using progressively shorter forecasting windows (Tab. \ref{tab2}), using the U-Net 3D as backbone architecture.
Although the full dataset includes temporal sequences spanning ten days (four days before to five days after ignition), these supplementary models are limited to post-ignition data up to one, two, three, four, or five days after the fire started.
Each model retained the same architecture as the primary 10-day model, only differing in the amount of post-ignition information available during training. 
All models were tested on the same independent test set used for the main performance comparison to ensure consistency and comparability across temporal configurations.

\begin{table}[h!]
\centering
\caption{\textbf{Evaluation of model performance across different forecasting horizons of 5, 4, 3, 2, and 1 days. Results illustrate how predictive accuracy varies with the length of the forecasting window, with higher performance metrics highlighted in bold.}}
\label{tab2}
\setlength{\tabcolsep}{4pt}
\renewcommand{\arraystretch}{1.2}
\begin{tabular}{cccccc}
\hline
\textbf{Metric} & \textbf{U-Net3D} & \textbf{U-Net3D} & \textbf{U-Net3D} & \textbf{U-Net3D} & \textbf{U-Net3D} \\
\textbf{\%}    & \textbf{10 days} & \textbf{9 days} & \textbf{8 days} & \textbf{7 days} & \textbf{6 days} \\
\hline
Dice Score & \textbf{53.6} & 51.2 & 49.4 & 48.7 & 47.5 \\
IoU        & \textbf{36.6} & 34.4 & 32.8 & 32.2 & 31.1 \\
Precision  & \textbf{59.6} & 56.6 & 55.0 & 53.6 & 53.6 \\
Recall     & \textbf{48.8} & 46.7 & 44.9 & 44.6 & 42.5 \\
\hline
\end{tabular}
\end{table}

A consistent trend observed across these experiments is that model performance declines as the number of post-ignition days included in the input decreases.
The model utilizing the full 10-day temporal context---including data from five days post-ignition---achieves up to a 6.1\% higher Dice coefficient compared to variants using fewer forecasting days.
This highlights the substantial benefit of incorporating post-ignition temporal data to enhance wildfire spread predictions.

These findings are consistent with previous research and emphasize the critical role of temporal dynamics in modeling fire behavior. 
Wildfires typically evolve over multiple days, particularly in larger and more complex incidents, where factors such as shifting wind conditions and interactions with topography play a significant role in determining the fire’s progression. 
Limiting the temporal context, particularly by excluding key post-ignition periods, restricts the model’s ability to capture these evolving dynamics, ultimately reducing predictive accuracy.

\subsection{Results per Fire Cluster}
To gain a deeper insight into how model performance varies with fire size, evaluation results are reported per each separate fire size cluster, as presented in Tab. \ref{tab3}, for the best-performing model (10-day U-Net 3D).
The results show that the model performs well for smaller-scale events, achieving a Dice Score of approximately 59\% within the first three clusters, which represent fires up to 5000 hectares. 
A slight drop in performance of around 2.6\%, is observed in the fourth cluster, covering fires up to 8500 hectares, while for fire events exceeding this threshold, model accuracy steadily declines. 

This downward trend correlates with the increasingly limited availability of training samples in the larger fire clusters, as evident from the number of samples per cluster in Tab. \ref{tab3}. 
The scarcity of examples for large-scale fires likely restricts the model's ability to generalize effectively in these cases, contributing to the observed decrease in segmentation accuracy. 
These results highlight once again the challenge posed by class imbalance in wildfire-related applications.

\begin{table}[h!]
\centering
\caption{\textbf{Average performance metrics per fire cluster using the 10-day U-Net 3D model, computed from test samples within each cluster.}}
\label{tab3}
\setlength{\tabcolsep}{2.5pt}
\renewcommand{\arraystretch}{1.2}
\begin{tabular}{ccccccccc}
\hline
\textbf{Metric} & \textbf{250} & \textbf{1500} & \textbf{3000} & \textbf{5000} & \textbf{8500} & \textbf{14000} & \textbf{20000} & \textbf{>} \\
\textbf{\%}    & \textbf{1500} & \textbf{3000} & \textbf{5000} & \textbf{8500} & \textbf{14000} & \textbf{20000} & \textbf{35000} & \textbf{35000} \\

\hline
 &  &  &  Number & of &  Samples &  &  &  \\
Number & 451 & 298 & 161 & 107 & 40 &  22 & 18 & 4 \\
\hline
Dice Score & 58.6 & 58.4 & 58.0 & 55.4 & 50.9 & 57.9 & 30.7 & 15.5 \\
IoU        & 41.5 & 41.3 & 40.9 & 38.4 & 34.1 & 40.8 & 18.1 & 8.40 \\
Precision  & 47.6 & 53.7 & 61.4 & 63.9 & 69.9 & 82.6 & 84.1 & 95.5 \\
Recall     & 76.3 & 64.1 & 55.0 & 48.9 & 40.0 & 44.6 & 18.8 & 8.43 \\
\hline
\end{tabular}
\end{table}


\Figure[h!](topskip=0pt, botskip=0pt, midskip=0pt)[width=1.0\linewidth]{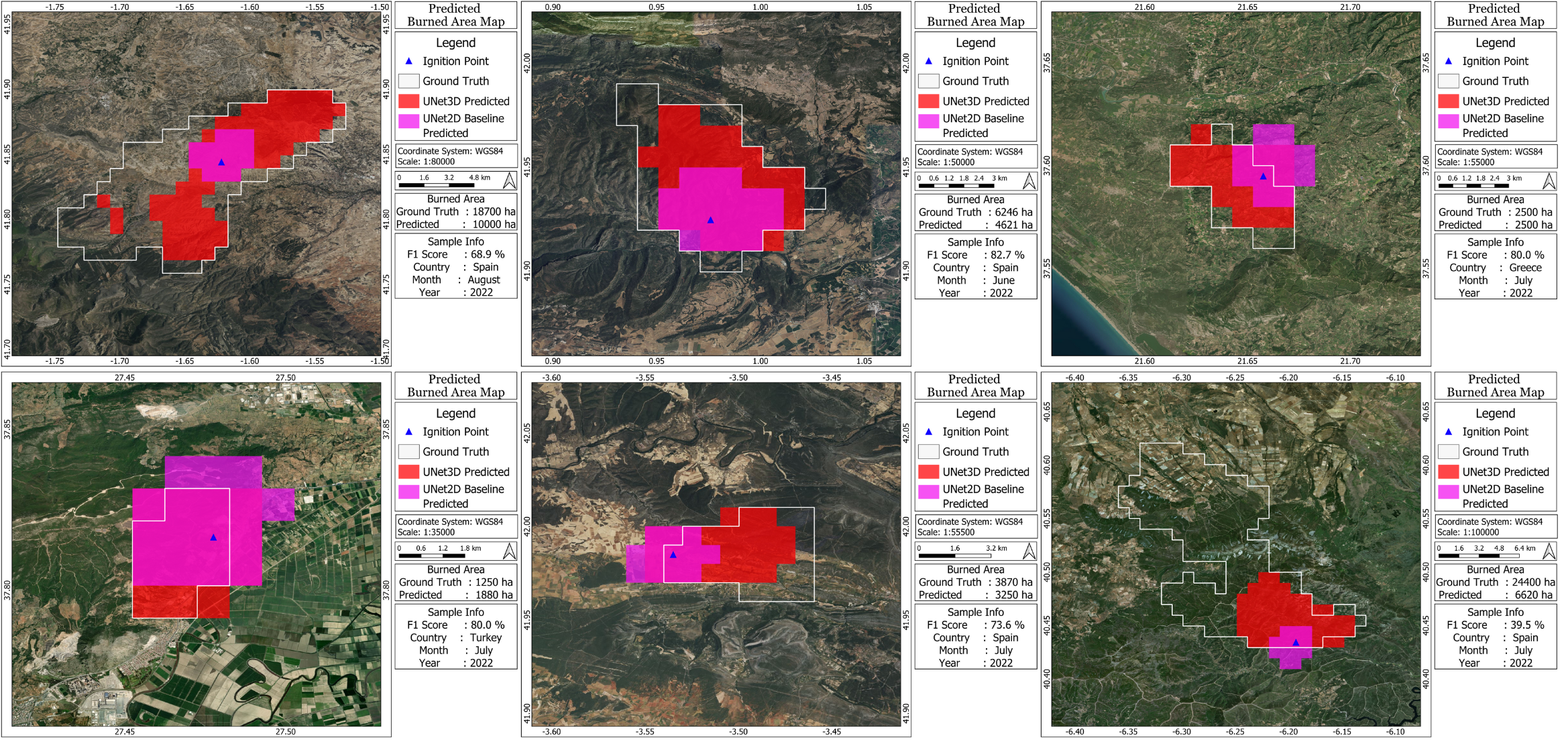}
{ \textbf{Visualizations of predicted burned area maps from the test set, comparing the best-performing 10-day U-Net3D model with the ground truth and the baseline U-Net2D model. The results demonstrate that the 10-day U-Net3D model consistently outperforms the baseline, accurately capturing the extent and spread of burned areas. It effectively predicts the majority of affected regions in both large and small fire events, with well-contained and correctly directed spread patterns.}\label{maps_test}}

\subsection{Qualitative Evaluation} 
To facilitate a qualitative evaluation of the models, Fig. \ref{maps_test} presents representative prediction maps from the test set, showcasing the outputs of the baseline model (ignition-day U-Net2D) and the best-performing model (10-day U-Net3D), alongside the ground truth fire extent.
In these visualizations, the baseline model’s predicted burned area is depicted in purple, the 10-day U-Net3D model’s prediction in red, and the ground truth fire boundaries in white.

The 10-day U-Net3D model demonstrates superior performance over the baseline for both large and small fire events.
The baseline model, trained solely on ignition-day data, predicts a more uniform radial spread around the ignition, failing to accurately capture the anisotropic propagation of the fires. 
In contrast, the 10-day U-Net3D model, with access to multi-day temporal data---including the five days following ignition---is better equipped to capture the evolving influence of dynamic environmental conditions, such as changes in wind direction and speed.

Furthermore, the 10-day U-Net3D model also shows strong capability in accurately delineating the ground truth fire extent. 
Across most test samples, the predicted burned area is well contained, showing only small overestimation or underestimation of the fire boundaries. 
The spatial progression of the fire spread is also consistently aligned with the observed direction of the ground truth events, indicating once again that the model effectively captures the dynamic behavior of fire propagation. 



The last image (second row, third column) illustrates a challenging case in which both models fail to accurately predict the final burned area. 
Although the general direction of spread is partially captured, neither model successfully forecasts the final size or precise spatial extent of the fire. 
However, even in this case, the best-performing model has captured the direction of spread.

These visualizations underscore again that the 10-day U-Net3D model achieved the most favorable results, excelling in both quantitative metrics and qualitative assessment of prediction maps. 
\section{Limitations}
While this study provides valuable insights into data-driven modeling of wildfire spread, several limitations should be considered when interpreting the results.

A key limitation lies in the temporal forecasting window, which includes data up to five days post-ignition.
However, smaller fire events often extinguish within one or two days, while very large fires may continue burning beyond the five-day window. 
Nonetheless, the five-day window was selected to align with the temporal horizon of operational meteorological forecasts, which typically offer limited reliability beyond that range. 

Another temporal constraint arises from the use of actual meteorological observations rather than forecast data, due to the unavailability of historical weather forecasts. 
While this choice supports accurate retrospective analysis, it introduces a distribution shift from real-world operational conditions, where models would rely on forecasted weather inputs.

The spatial resolution of 1 km $\times$ 1 km, determined by Mesogeos, limits the model’s ability to capture finer-scale fire dynamics that occur at sub-kilometer levels.
Although this resolution is suitable for many operational applications, higher-resolution data (e.g., 300 m) are available for certain input variables. like topography, and could potentially enhance model performance.




\section{Conclusion}
Wildfires represent a persistent and escalating threat to both human populations and natural ecosystems, underscoring the need for effective predictive tools.  
This study demonstrated the feasibility of forecasting the final burned extent of wildfire events from the day of ignition using DL methods.

We introduced a curated, ML-ready dataset and developed DL models that proved effective in capturing the complex spatio-temporal patterns governing fire behavior.
Our findings demonstrated that a purely data-driven DL approach can successfully model wildfire dynamics.
Furthermore, we showed that incorporating a more extended temporal context, specifically data spanning up to five days post-ignition, substantially improved predictive performance, highlighting the importance of temporal dynamics in modeling fire spread.
Finally, the proposed models maintained satisfactory performance across all fire size categories, demonstrating promising skill, both quantitative and qualitative, even on larger events.


Future efforts could consider integrating additional wildfire-related variables---particularly detailed vegetation maps or land cover variables--- to improve model understanding of fuel availability and flammability.
Moreover, oversampling of large fire events could be used to address the imbalance problem of large fires. 
Finally, exploring advanced architectures such as ConvLSTMs, which are designed to jointly capture spatial and temporal dependencies, may lead to more accurate modeling of fire spread.

\section*{Acknowledgment}
This work was funded by the project: Small-Satellites (Measure ID 16855), implemented by the Hellenic Ministry of Digital Governance with the European Space Agency (ESA) Assistance in the Management and Implementation. The project is part of the National Recovery and Resilience Plan ‘Greece 2.0’, which is funded by the Recovery and Resilience Facility (RRF), core programme of the European Union-NextGenerationEU.

\bibliographystyle{IEEEtran}
\bibliography{references}

\section{appendix}
Table \ref{tab4} lists all the variables used, along with their initial spatial and temporal resolutions, categorized accordingly.

\begin{table}[h!]
\caption{\textbf{Variables used in Study. The variables are categorized into four sections, meteorological variables, vegetation maps, land cover variables and thermal anomalies.}}
\label{tab4}
\setlength{\tabcolsep}{2pt}
\begin{tabular}{p{100pt}cc}
\hline
\multicolumn{1}{c}{\textbf{Variable}} & \multicolumn{1}{c}{\textbf{Spatial Resolution}} & \multicolumn{1}{c}{\textbf{Temporal Resolution}} \\

\hline
& \centering \emph{Meteorological} & \\
\hline
Max Temperature &  9x9 km &  1 day \\
U Component of Wind & 9x9 km & 1 day \\
V Component of Wind & 9x9 km & 1 day \\
Max Dewpoint Temperature & 9x9 km & 1 day \\
Max Surface Pressure & 9x9 km & 1 day \\
Min Relative Humidity & 9x9 km & 1 day \\
Total Precipitation & 9x9 km & 1 day \\
Mean Surface Solar Radiation & 9x9 km & 1 day \\
Day Land Surface Temperature & 1x1 km & 1 day \\
Night Land Surface Temperature & 1x1 km & 1 day \\

\hline
& \centering \emph{Vegetation Maps} & \\
\hline
Vegetation Index (NDVI) & 0.5x0.5 km & 16 days \\
Leaf Area Index (LAI) & 0.5x0.5 km & 8 days \\
Soil moisture & 5x5 km & 10 days \\

\hline
& \centering \emph{Land Cover} & \\
\hline
Fraction of Agriculture & 0.3x0.3 km & - \\
Fraction of Forest & 0.3x0.3 km & - \\
Fraction of Grassland & 0.3x0.3 km & - \\
Fraction of Settlements & 0.3x0.3 km & - \\
Fraction of Shrubland & 0.3x0.3 km & - \\
Fraction of Sparse Vegetation & 0.3x0.3 km & - \\
Fraction of Water Bodies & 0.3x0.3 km & - \\
Fraction of Wetland & 0.3x0.3 km & - \\

\hline
& \centering \emph{Topography} & \\
\hline
Elevation & 0.03x0.03 km & - \\
Slope & 0.03x0.03 km & - \\
Aspect & 0.03x0.03 km & - \\
Curvature & 0.03x0.03 km & - \\

\hline
& \centering \emph{Thermal Anomalies} & \\
\hline
Burned Areas & 1x1 km & 1 day \\
Ignition Points & 1x1 km & 1 day \\

\hline
\end{tabular}
\label{tab4}
\end{table}

\EOD

\end{document}